\RequirePackage{amsmath}
\documentclass[runningheads]{llncs}
\newcommand{\methodName}{DMCA }
\usepackage[T1]{fontenc}
\usepackage{booktabs}
\usepackage{epstopdf}
\usepackage{graphicx}
\usepackage{svg}  

\begin{document}
\title{Dual-Modality Representation Learning for Molecular Property Prediction}
%
%
\author{Anyin Zhao\inst{1}\orcidID{0009-0008-3785-8568} \and
Zuquan Chen\inst{1}\orcidID{0009-0006-0324-3448} \and
Zhengyu Fang\inst{1}\orcidID{0000-0001-7835-0543} \and
Xiaoge Zhang\inst{1}\orcidID{0009-0007-9839-5854} \and
Jing Li\inst{1,*}\orcidID{0000-0003-1160-6959}}
\authorrunning{A. Zhao et al.}
%
\institute{Case Western Reserve University, Cleveland OH 44106, USA 
\email{jingli@case.edu}\\
}
\maketitle              
\thispagestyle{empty}
\begin{abstract}
Molecular property prediction has attracted substantial attention recently. Accurate prediction of drug properties relies heavily on effective molecular representations. The structures of chemical compounds are commonly represented as graphs or SMILES sequences. 
Recent advances in learning drug properties commonly employ Graph Neural Networks (GNNs) based on the graph representation. For the SMILES representation, Transformer-based architectures have been adopted by treating each SMILES string as a sequence of tokens. Because each representation has its own advantages and disadvantages, combining both representations in learning drug properties is a promising direction. We propose a method named Dual-Modality Cross-Attention (DMCA) that can effectively combine the strengths of two representations by employing the cross-attention mechanism. DMCA was evaluated across eight datasets including both classification and regression tasks. Results show that our method achieves the best overall performance, highlighting its effectiveness in leveraging the complementary information from both graph and SMILES modalities.

\keywords {Molecular property prediction \and Dual-modality learning \and Cross-attention learning.}
\end{abstract}
\newpage
\setcounter{page}{1}
\section{Introduction}
It is well known that traditional drug discovery is costly, time-consuming, and with high failure rates~\cite{mullard2014new}. To streamline the process of drug discovery and mitigate resource-intensive laboratory work, significant research has been dedicated to the development of computational methods for drug property prediction. 
Recently, deep learning technologies have achieved remarkable success across various domains~\cite{zemouri2019deep,choudhary2022recent,yue2024phase,jumper2021highly}. In this paper, we focus on deep-learning approaches for molecular property prediction, 
the primary goal of which is to generate hypotheses based on the prediction results, and to enable researchers to prioritize and validate those hypotheses~\cite{wang2022molecular,zhu2021dual,guo2021multilingual,hu2019strategies}. 

To predict molecular properties accurately,  it is essential for all deep learning approaches to first establish strong representations of molecules. Molecules and chemical compounds can be represented differently. One intuitive and commonly used approach is to use a 2D graph to represent the chemical structure of a molecule, where nodes and edges represent atoms and bonds, respectively. Naturally, graph neural networks (GNNs)~\cite{kipf2016semi} have been adopted for handling graph representations of molecules for property prediction. Alternatively, the same molecule can be represented as a SMILES~\cite{weininger1988smiles} sequence, which can be generated based on the chemical structure of a molecular by applying a set of predefined rules. Because SMILES sequences consist of atom names and special symbol tokens that mimic sequences of words in natural language text, it is not surprising that natural language processing methods such as Transformers have been adopted to process the SMILES format of drug molecules~\cite{vaswani2017attention}. 

Both molecular representations and their corresponding processing methods have strengths and limitations. Recent works~\cite{gilmer2017neural,yang2019analyzing} have adopted GNNs for molecular property prediction and achieved promising results. GNNs process the molecules as graphs and iteratively update node representations based on information from neighboring nodes, which can effectively capture local information and identify functional groups. However, they struggle to integrate information from distant nodes, limiting their ability to learn interactions between functional groups~\cite{li2018deeper,xu2018powerful}. Given the success of Transformer-based models~\cite{vaswani2017attention} in natural language processing, recent studies have adapted similar models for drug property prediction by treating SMILES sequences as text~\cite{chithrananda2020chemberta,wang2019smiles}. These approaches usually perform well by leveraging pre-training on large molecule datasets and can capture global sequence-level information. However, they struggle to directly encode topological features, such as chemical rings, in their models.

Since these two representations 
offer complementary information, combining them by utilizing some of the most recent advances in multi-modality learning could possibly enhance the prediction performance. Recently, Transformer-based multi-modality learning has attracted great attention in the Machine Learning community and has shown superior performance in many application domains (e.g., references in a recent review paper~\cite{Xu2023}). Leveraging such success, in this work, we propose a novel transformer-based network structure for multi-modality learning of molecular representations. Our hierarchical network model takes two data streams of two different modalities of drug molecules as inputs. A GNN-based encoder and a transformer-based encoder first learn the representations of molecules based on their graph and SMILES representations, respectively. A Transformer based multimodal encoder then fuses the features learned from different modalities through cross-modality attentions. The output can then feed to a multi-layer perceptron (MLP) network that can be used directly to predict molecular properties. We applied our Dual-Modality Cross-Attention (DMCA) model on eight datasets from the MoleculeNet Benchmark data~\cite{wu2018moleculenet}. Results show that DMCA achieves the best overall performance, highlighting its effectiveness in leveraging the complementary information from both graph and SMILES modalities.

\section{Related work}
Recent research on molecular property prediction includes graph-based methods, SMILES-based methods, and multi-modality methods. We briefly review some of the most recent developments for each type of the methods and many of which will serve as the baseline approaches in our experiments.\\ 
%
\textbf{GNN-based approaches for molecular representation} With molecules intuitively represented as graphs, GNNs offer a natural framework for handling and analyzing molecular data. This synergy has sparked extensive research, positioning GNNs at the forefront of innovation in molecular property prediction. GNNs allow nodes to aggregate information through their edges, creating comprehensive graph representations. Furthermore, by combining graph structures with neural networks, GNNs can readily handle both classification and regression tasks. Therefore, many innovations in GNNs  (e.g., \cite{gilmer2017neural,yang2019analyzing}) focus on graph representation learning, rather than specific prediction tasks. Similar to other deep learning frameworks, in order to obtain a more robust representation, many GNN models have started to explore the strategy of pre-training. 
For example, Hu et al.~\cite{hu2019strategies} adapted a pre-training strategy at both the individual node and entire graph levels with attribute masking to learn better local features. MolCLR~\cite{wang2022molecular} used the contrastive learning strategy based on the positive and negative pairs created by different graph augmentation methods. GROVER\cite{rong2020self} relied on self-supervised pretraining on unlabeled graphs and integrated Message Passing Networks into the Transformer-style architecture. As a geometry-based GNN, GEM\cite{fang2022geometry} captured molecular spatial knowledge pretrained via geometry-level self-learning methods. GraphMVP \cite{liu2021pre} incorporated 3D information to augment its 2D graph representation by self-supervised learning.
While GNNs can easily capture local connectivity information from the graphs by integrating information from surrounding nodes when updating node features, they often struggle to model the global features of molecules. In order for GNNs to learn more complex features, one often has to increase the number of layers in the networks. However, increasing the number of layers can lead to  over-smoothing~\cite{li2018deeper}, where the representations of nodes can become very similar.\\
\textbf{SMILES-based approaches for molecular representation} Previously, various RNNs have been applied to SMILES to extract molecular representations. Recently, Transformer architectures \cite{devlin2018bert,vaswani2017attention} have demonstrated great success in natural language processing and other applications using sequential data. Because SMILES can be considered as a language utilized to depict molecular structures, researchers consider Transformer-based architectures as a promising option for learning molecular representation \cite{chithrananda2020chemberta,wang2019smiles,xue2020x,li2021mol}.
Many researchers applied different pre-training strategies on molecules based on Transformers. ST\cite{honda2019smiles} was an encoder-decoder network based on Transformer, which was pretrained on unlabeled SMILES. Similar to the vanilla Transformer, SMILES-BERT \cite{wang2019smiles} and ChemBERTa \cite{chithrananda2020chemberta} utilized masked language modeling for pre-training on molecules. Fabian et al. \cite{fabian2020molecular} introduced additional tasks to SMILES-based molecule pre-training, including predictions on the equivalence of two different SMILES strings.
Similarly, X-Mol \cite{xue2020x} was pre-trained by generating a valid and equivalent SMILES representation from an input SMILES representation of the same molecule. Mol-BERT \cite{li2021mol} designed a feature extractor to transform SMILES into an atom identifier and adopted the masking technique for pre-training to obtain better molecular sub-structural information. Recently, some research applied message passing (MP) operations on SMILES. MPAD \cite{jo2020message} adopted an MP-based attention network to process SMILES to achieve molecular representation. Although these models provided insights for further research, their performance may not always be satisfactory due to the inherent limitation of the SMILES format. While SMILES can easily capture global information, they cannot easily extract structural information such as rings in a molecule.\\
\textbf{Multi-modality learning for molecular representation} Multi-modality learning has shown remarkable success in many different domains~\cite{bao2021vlmo,kim2021vilt,li2021align,jia2021scaling,radford2021learning}. The commonly used modalities include language/text, graph, video/image, and audio. Generally speaking, multi-modality learning can be characterized as Joint representation learning and Coordinated representation learning \cite{baltruvsaitis2018multimodal}. Joint representation learning projects all modalities on the same space via various strategies such as concatenation or dot product. Coordinated representation learning coordinates different modalities via distance 
or structure constraints, 
enabling cross-modality interaction learning through a contrastive loss.
For molecular property prediction, the two most commonly used modalities are SMILES and chemical graph structures. For example, similar to our work, both DVMP \cite{zhu2021dual} and MMSG \cite{wu2023molecular} utilized these two modalities; but they differed in their learning strategies. DVMP \cite{zhu2021dual} adapted the Coordinated Representation Learning strategy and tried to generate a more robust representation by maximizing the consistency between the outputs from the Transformer for SMILES and the GNN branch for 2D chemical graphs. MMSG\cite{wu2023molecular} followed the Joint Representation learning strategy and enhanced the interactions between the two modalities by replacing attention bias in Transformers handling encoded SMILES with bond-level graphs. Different from the two approaches, MSSGAT\cite{ye2022molecular} designed a framework including graph attention convolutional (GAC) blocks and deep neural network (DNN) blocks to process three types of features: raw molecular graphs, molecular structural features via tree decomposition, and Extended-Connectivity FingerPrints (ECFP), followed by the concatenation of the three embeddings. Our own model falls into this category. One critical distinction is that we actually use cross-attention to jointly learn the interactions among the embeddings from different modalities.   
%
%
\begin{figure*}[t]
	\centering
	\includegraphics[width=\linewidth]{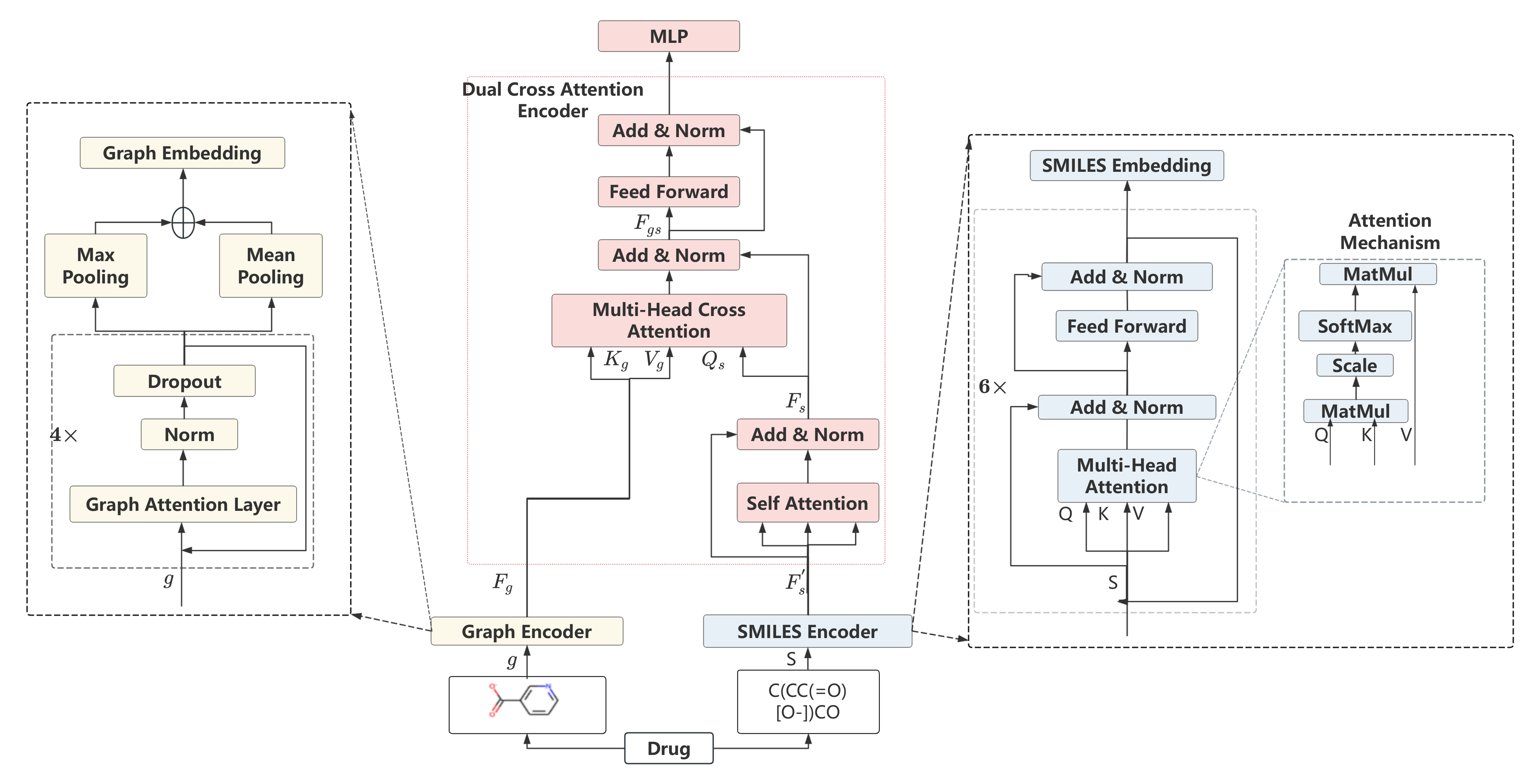}
	\caption{Workflow of the proposed \methodName model. The left side is the Graph Encoder; the right side is the SMILES Encoder; and the middle part is the cross-attention module.}
	\label{fig:model_workflow}
\end{figure*}
\section{Methods}
In this section, we introduce our method DMCA, which is a hierarchical deep neural network designed for joint SMILES-graph learning. The key idea of \methodName is to improve the performance of molecular representation learning through transformer-based multi-modality learning. In particular, \methodName explicitly integrates information from the dual modalities via a cross-attention mechanism to generate the final representation. As shown in Fig. 1, the overall architecture of \methodName consists of three components: (i) a GNN branch that takes molecular graphs as input, which aims to capture the local structure information of molecules; (ii)  a Transformer branch that takes SMILES strings as input, which focuses on capturing the global information of molecules; and (iii) a cross-attention Transformer that jointly learns the interactions from the embeddings of these two branches/modalities.
\vspace{-0.25cm}
\subsection{Graph Encoder}
Our graph encoder is a stack of four Graph Attention Network (GAT) layers as its backbone \cite{velivckovic2017graph}, with graph normalization \cite{ioffe2015batch} and non-linear activation followed at each layer. More specifically, each GAT layer takes a set of node features $h$ as its input, where \(h = \{h_1, h_2, ..., h_N\}, h_i \in R^F\), $N$ is the number of nodes and $F$ is the number of features for each node. It generates a set of new node features \(h^{'}\) with potentially different feature cardinality, by considering contributions from all neighbor nodes. For each node $i$ and each of its neighbors $j$, the GAT layer applies a shared attention mechanism \(a\), which is a single-layer feed-forward neural network with a weight matrix \(W^a\) followed by a LeakyReLU activation function, to compute the importance of node \(j\) to node \(i\) as
\[
    e_{ij} = LeakyReLU((W^a)^T[Wh_i || Wh_j]),
\]
where \( W \in R^{F^{'} \times F} \), \( W^a \in R^{2F^{'}} \), $F^{'}$ is dimensionality of the new feature set,  and \( || \) represents the concatenation operation. The importance score is further normalized via a softmax function to obtain the normalized weight
\[
    \alpha_{ij} = \text{softmax}_j(e_{ij}).
\]
The contributions of all neighboring nodes are summed up via a weighted linear combination of the node features, followed by a nonlinear transformation \( \sigma \), layer normalization, and a dropout strategy, to obtain new feature \( h^{'}_i \) via the formula:
\[
        h^{'}_i = \text{D}(\text{LN}(\sigma( \sum_{j\in N_i}\alpha_{ij} W h_j))),
\]
where D is dropout layer and LN is the normalization layer.

This operation is iterated for four times. After the last GAT layer, we use both mean pooling and max pooling and concatenate them to get the final graph embedding. We utilize a relatively simple graph encoder because our main focus is to investigate whether the proposed dual-modality learning via cross-attention can significantly improve performance. For node features, we adopt the standard feature set for chemical molecules and use numerical values for simplicity\cite{xu2023molecular}. A summary of these features is provided in Table \ref{tab:graph_node_feature_table}. 
\vspace{-0.25cm}
\subsection{SMILES Encoder}
SMILES is the most widely used linear representation for describing chemical structures~\cite{weininger1988smiles}. Transformer-based systems treat each SMILES as a sequence of characters and words, and need a tokenizer to recognize individual words. We adopt the tokenizer from an earlier work~\cite{chithrananda2020chemberta}, which is based on Byte-Pair Encoder (BPE) from HuggingFace tokenizers library~\cite{wolf2020transformers}, and is a hybrid between character-level and word-level representations. The size of the vocabulary is 767 and the maximum sequence length is 512, which is sufficient for most SMILES sequences. 
%
Transformer architectures excel at capturing global features from textual data. Building on this capability, we leverage the attention mechanism of the Transformer to model global relationships within molecular structures. The Transformer encoder comprises two sub-layers: a multi-head attention layer and a fully-connected feed-forward layer. Each sub-layer is followed by a residual connection and layer normalization to normalize input values for all neurons in the same layer \cite{vaswani2017attention}, as illustrated in Fig \ref{fig:model_workflow}.

The multi-head attention mechanism involves performing self-attention multiple times with different weight values to learn various features. The computation process for the self-attention mechanism is summarized by the formula:
\[
\begin{array}{c}
    Q = SW^Q,  K = SW^K,  V = SW^V \\
    \text{Attention}(Q, K, V) = \text{softmax}\left(\frac{QK^\top}{\sqrt{d_k}}\right)V
\end{array}
\]
where \(S\) is the SMILES string for each molecule after tokenization, \(W^Q, W^K, W^V\) are the query, key, and value weight matrices, and \(d_k\) is the dimension of \(S\). By stacking this encoder multiple times, we obtain the final embedding \(F^{'}_s\) for subsequent fusion.  To improve the performance and reduce training time, we adopt a pretrained model ChemBERTa \cite{chithrananda2020chemberta}, which was pretrained via self-supervised learning on the Zinc dataset  \cite{irwin2005zinc} and consisted of six layers and twelve heads.
\vspace{-0.25cm}
\subsection{Dual-Modality Cross-Attention Encoder}
The structure of the Cross-Attention Encoder is shown in Fig \ref{fig:model_workflow}, the goal of which is to learn a joint representation of molecules based on the graph embedding and the SMILES embedding. The graph embedding from the GNN branch is denoted as \(F_G\). The embedding output from the Transformer encoder is denoted as \(F^{'}_s\). We first apply the self-attention mechanism on it, followed by a layer normalization and residual connection, to further learn the global relationship. The final embedding is denoted as \(F_s\). To fuse the two embeddings \(F_g\) and \(F_s\) via cross-attention,  we first partition them into \(h\) parts/heads and use \(F_s\) to calculate the query matrix and \(F_g\) to calculate the key and value matrices by multiplying the corresponding weight matrices: 
\[
    Q_{si} = F_s W^Q_{si}, K_{gi} = F_g W^K_{gi}, V_{gi} = F_g W^V_{gi}. \\
\]
The cross-attention is then applied for each attention head via a Softmax function: 
\[
    head_i = \text{CrossAttention}(Q_{si}, K_{gi}, V_{gi}) = \text{softmax}\left(\frac{Q_{si}(K_{gi})^T}{\sqrt{d_s}}\right)V_{gi}. 
\]
Finally, multi-head cross-attention is achieved via concatenation:
\[
    \text{MultiHeadCrossAttention}(Q_{s}, K_{g}, V_{g}) = \text{Concat}(head_1, head_2, ..., head_h) W^o, \\
\]
where \(W^o\) is a randomly generated projection matrix. We further add a residual connection layer with layer normalization to get \(F_{GS}\), followed by a feed-forward layer with residual connection and layer normalization to get the final joint representation. The joint representation can then be utilized for any downstream applications, 
for example, via a MLP layer at the last step.




\section{Experiments}
To assess the effectiveness of our proposed method, we performed multiple challenging classification and regression tasks based on the MoleculeNet Benchmark datasets \cite{wu2018moleculenet}. We use the Area under the Receiver Operating Characteristic Curve(AUC-ROC) and the Root Mean Square Error(RMSE) as evaluation metrics for classification and regression tasks, respectively. We also conduct ablation studies to assess the effectiveness of the cross-attention mechanism as a joint learning approach. 
\vspace{-0.25cm}
\subsection{Datasets}
The MoleculeNet Benchmark \cite{wu2018moleculenet} is a comprehensive online repository consisting of many different datasets. In this study, we select eight commonly used datasets spanning diverse sets of molecular properties.
The prediction tasks can be categorized into classification tasks and regression tasks. Five datasets, namely BBBP, BACE, ClinTox, HIV and SIDER, are used for classifications. Each of them features SMILES strings as molecular descriptors and one or multiple binary labels for molecular property(ies). Among them, BBBP, BACE, and HIV datasets only consist of a single-label. ClinTox has two labels, indicating whether a drug is FDA approved (\texttt{FDA\_APPROVED}) and its toxicity (\texttt{CT\_TOX}). While most previous work simply treated the ClinTox dataset as a dual-label prediction task without considering the relationship between these two labels, they actually have almost perfect correlations. An FDA approved drug (\texttt{FDA\_APPROVED=1}) most likely will and should pass clinical trials for toxicity (\texttt{CT\_TOX = 0}); and a toxic compound (\texttt{CT\_TOX = 1}) is very unlikely approved by FDA (\texttt{FDA\_APPROVED=0}). Indeed, these two combinations contribute to $98.8\%$ of all the entries (Fig. A1). Therefore, in our study, we simply treat it as a dataset with a single label. The SIDER dataset originally had 5880 different types of side effects. They are then grouped into 27 organ classes based on MedDRA classification \cite{kuhn2016sider}. The class labels afterward do not have much correlations anymore. We again consider each label independently and perform single-label prediction. The performance on this dataset is computed based on the mean AUC-ROC across all labels. The three datasets for the regression task include ESOL, FreeSolv, and Lipophilicity, each of which consists of SMILES strings as molecular descriptors and a quantitative molecular property. The length distribution of the SMILES strings from each dataset can be found in Fig. A1. Significant majority of them have a length less than 512, the parameter used for SMILES string tokenization. Graph representation of each molecule is obtained from SMILES using the RDKit package. Brief description of the datasets can be found in the Appendix.
\vspace{-0.25cm}
\subsection{Implementation}
We implemented the proposed approach, including the GNN branch, using the PyTorch library. 
The SMILES encoder is the same as the encoder part of ChemBERTa with an output dimension of 767. The dual-modality cross-attention component is constructed with 767-dimensional hidden units and employs 13 attention heads to capture diverse interactions and dependencies across the two modalities. We used the cross-entropy as the loss function and adopted the Adam Optimization method. The experiments were run on the HPC of the Ohio Super Computer Center. 
\subsection{Experiment design}
For each dataset, the training/testing split ratio is 0.8/0.2. Each task was independently run three times with random seeds and the means and standard deviations of the performance measure were recorded. 
We selected 10 baseline methods for comparison based on their reported performance and the alignment with our method's research direction, which included five graph-based methods (Hu\cite{hu2019strategies}, GROVER\cite{rong2020self}, MolCLR\cite{wang2022molecular}, GEM\cite{fang2022geometry}, GraphMVP \cite{liu2021pre}), two SMILES-based approaches (MPAD\cite{jo2020message} and ST\cite{honda2019smiles}), and three multi-modality learning approaches (DVMP\cite{zhu2021dual}, MSSGAT\cite{ye2022molecular} and MMSG\cite{wu2023molecular}). GROVER had two versions: GROVER\textsubscript{base} and GROVER\textsubscript{large}, and results from both versions were included. MPAD and MSSGAT lacked experiments on the regression tasks and are not included in the evaluation of regression tasks. In addition, we also performed ablation studies by considering each of the two branches alone (ChemBERTa and GAT) and included their results. All the results from the baselines (other than the ablation study) were obtained directly from their corresponding publications. We should clarify that previous researchers might have used different data splitting strategies.

\begin{figure}[t]
    \centering
    \includegraphics[width=\textwidth]{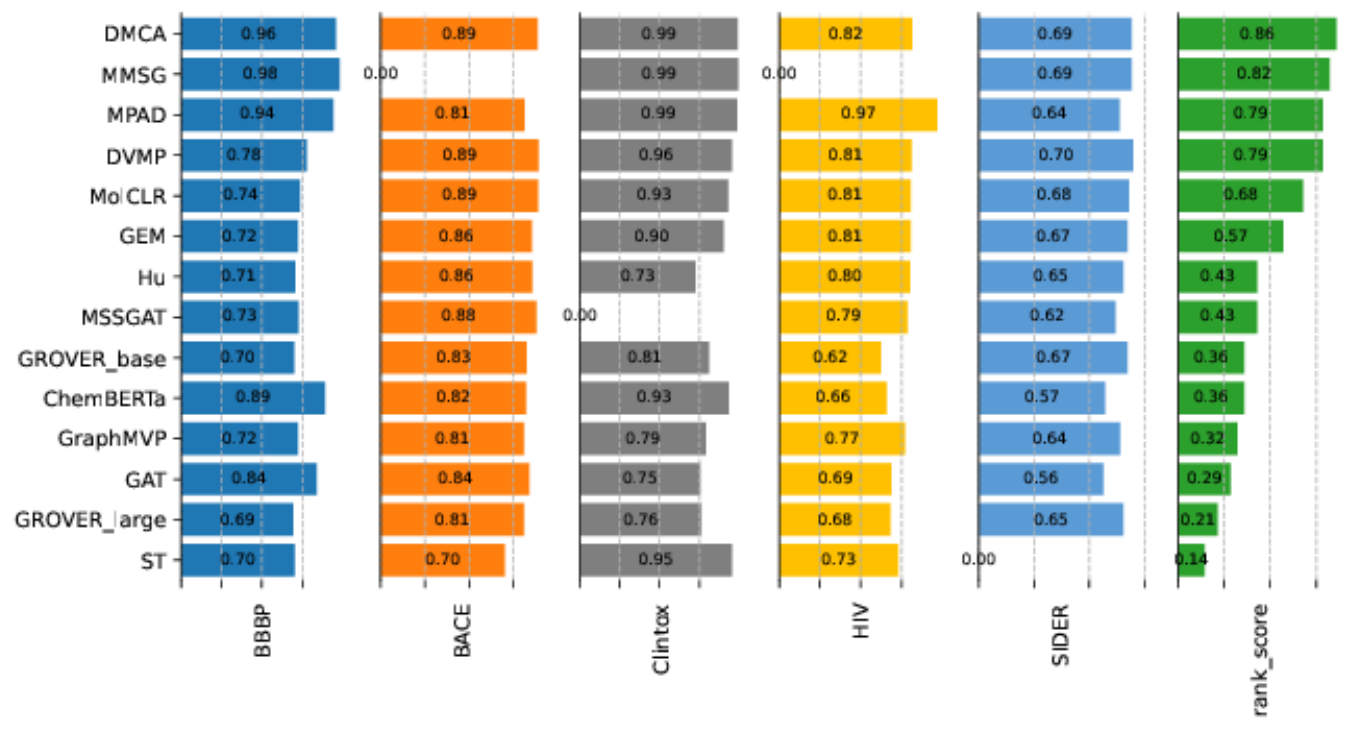}
    \caption{Results of the molecule classification task. The last column (in green) shows the rank score. The methods are ordered based on this rank score.}
    \label{fig:molecule_results}
\end{figure}

\section{Results}
For the classification tasks, not a single method outperforms all other approaches in terms of AUC. To have a fair comparison for all methods across all datasets, we create a new measure called rank score by following the procedure below. Let $m_{i,j}$ denote the AUC of method $i$ on dataset $j$. For each dataset $j$, let $m_j^{\text{max}}$ and $m_j^{\text{min}}$ denote the best and worst AUC out of all the methods. We first apply the min-max scaling to normalize the AUC across all methods for each dataset $j$: 
\[
m_{i,j}^{\text{norm}} = \frac{m_{i,j} - m_j^{\text{min}}}{m_j^{\text{max}} - m_j^{\text{min}}}
\]
To derive an overall performance score for each method across all datasets, we propose a rank score by taking the median of the normalized AUC scores of each method across all datasets:
\[
\text{Rank Score}_i = \text{Median}(m_{i,1}^{\text{norm}},
m_{i,2}^{\text{norm}},
\dots, 
m_{i,n}^{\text{norm}})
\]
where $n$ is the total number of datasets. Finally, the rank score is utilized to rank all the methods, reflecting their overall performance (Fig.~\ref{fig:molecule_results}).


The results show that our proposed method achieves the best overall performance across all the datasets for the classification task (Fig. \ref{fig:molecule_results}). Among the top five methods, three of them are multi-modality-based approaches (DMCA, MMSG, and DVMP), one of them is a Transformer-based approach using SMILES as input (MPAD), and one of them is a GNN-based approach (MolCLR). This illustrates that consistent with the general belief, multi-modality approaches have some advantages over single-modality approaches. In addition to our approach, the multi-modality model MMSG also performed very well on the datasets that it had the results. However, it was somewhat penalized because it did not provide results on two of the datasets. We included five GNN-based approaches, but the best one from the five (MolCLR) only ranked fifth, indicating some inherent difficulties these methods faced.     

\begin{table}[t]
\caption{Results of mean and std of RMSE for the regression tasks in MoleculeNet benchmark.\cite{wu2018moleculenet}}
\centering
\begin{tabular}{@{}lllll@{}}
\toprule  
& ESOL & FreeSolv & Lipophilicity & Average \\
\midrule
ST \textsuperscript{\cite{honda2019smiles}} & $0.72 \pm N/A$ & $1.65 \pm N/A$ & $0.921 \pm N/A$ & 1.097 \\
Hu \textsuperscript{\cite{hu2019strategies}} & $1.100 \pm 0.006$ & $2.764 \pm 0.002$ & $0.739 \pm 0.003$ & 1.534 \\
GROVER\textsubscript{base}\textsuperscript{\cite{rong2020self}} & $0.983 \pm 0.090$ & $2.176 \pm 0.052$ & $0.817 \pm 0.008$ & 1.325 \\
GROVER\textsubscript{large}\textsuperscript{\cite{rong2020self}} & $0.895 \pm 0.017$ & $2.272 \pm 0.051$ & $0.823 \pm 0.010$ & 1.330 \\
MolCLR \textsuperscript{\cite{wang2022molecular}} & $1.11 \pm 0.01$ & $2.20 \pm 0.20$ & $ \mathit{0.65 \pm 0.08} $ & 1.320 \\
GEM \textsuperscript{\cite{fang2022geometry}} & $0.798 \pm 0.028$ & $1.877 \pm 0.024$ & $0.660 \pm 0.008$ & 1.112 \\
GraphMVP \textsuperscript{\cite{liu2021pre}} & $1.029 \pm N/A$ & $-$ & $0.681 \pm N/A$ & 0.855 \\
DVMP \textsuperscript{\cite{zhu2021dual}} & $0.817 \pm 0.024$ & $1.952 \pm 0.061$ & $0.653 \pm 0.002$ & 1.141 \\
MMSG \textsuperscript{\cite{wu2023molecular}} & $0.495 \pm 0.017$ & $\mathbf{0.712 \pm 0.087}$ & $\mathbf{0.538 \pm 0.007}$ & $\mathbf{0.582}$ \\
\midrule
ChemBERTa & $ \mathit{0.439 \pm 0.031 } $ & $1.333 \pm 0.052$ & $0.748 \pm 0.137$ & 0.840 \\
GAT & $0.921 \pm 0.049$ & $1.577 \pm 0.221$ & $0.710 \pm 0.007$ & 1.069 \\
DMCA & $\mathbf{0.432 \pm 0.037}$ & $ \mathit{1.321 \pm 0.061 } $ & $0.694 \pm 0.001$ & $\mathit{0.816}$ \\
\bottomrule
\end{tabular}
\label{table:regression result}
\end{table}

The results for the regression tasks are presented in Table \ref{table:regression result}. Two of the baselines (MPAD and MSSGAT) did not provide results on the three datasets and were excluded from the comparison. For the ESOL dataset, our model DMCA performed significantly better than all the baseline approaches. For the FreeSolv and Lipophilicity datasets, MMSG achieved the best results. DMCA ranked second on the FreeSolv dataset and MolCLR ranked second on the Lipophilicity dataset. When we computed the average RMSE of each method across all three datasets, MMSG ranked first and DMCA ranked second.

Overall, DMCA performed very well on all datasets. In addition, MMSG also performed well on all datasets where it provided results. The two methods share some commonalities but also have significant distinctions. MMSG extracts bond information from the graph branch as the attention bias, which plays a crucial role in improving its performance. On the other hand, we intentionally include a very simple model with a small number of layers for the GNN branch so we have a light-weighted model. Furthermore, we adopt a pre-trained model for the SMILES brach and do not require additional pre-training. Combining both strategies enables us to have a very efficient multi-modality model. 

\textbf{Ablation Study} In addition to the comparison with the baselines, we also perform ablation studies by assessing the performance of the two distinct branches alone (ChemBERTa and GAT) across all datasets. The ChemBERTa model is a pretrained RoBertA model based on SMILES, which is the branch used to obtain the text embedding in the \methodName model. We augment it with an MLP for downstream tasks and evaluate its performance accordingly. Similarly, the GAT model represents the graph-based branch of the \methodName model, responsible for generating graph embeddings.

We observe that across all datasets (Fig. \ref{fig:molecule_results} and Table \ref{table:regression result}), the performance of the dual-modality model \methodName outperformed both the ChemBERTa model and GAT model, clearly demonstrating the effectiveness of the cross-attention mechanism in combining information from the two complementary modalities. 
In addition, when comparing the ChemBERTa model with the GAT model, in most cases, ChemBERTa performs better than GAT, which is not surprising given the simple structure we used for GAT. But in some cases (e.g., Lipophilicity), GAT performed better than ChemBERTa. 
This highlights the significance of considering multiple modalities of molecular features and their interactions. Different molecular properties may depend on distinct types of features, necessitating a comprehensive approach, like our \methodName model, that incorporates diverse feature representations to achieve better performance.


\vspace{-0.25cm}
\section{Conclusion}
In this paper, we proposed a novel dual-modality representation learning method for molecular property prediction. By utilizing the cross-attention mechanism, the method can easily combine and enhance the learned representations from different modalities. We performed comprehensive experiments on eight diverse molecular datasets and compared the performance with ten baseline approaches. Results show that DMCA achieved the best overall performance on the classification tasks and ranked second on the regression tasks, highlighting its potential for a wide range of applications.


\newpage
%
%
%
\bibliographystyle{splncs04}
\bibliography{mybib}
%





\newpage
\setcounter{table}{0}
\setcounter{figure}{0}
\renewcommand{\thetable}{A\arabic{table}}
\renewcommand{\thefigure}{A\arabic{figure}}
\section*{Appendix}
\subsection*{A1. Summery of node features}
\begin{table}[ht]
\caption{Node features in graph representation}
\centering
\begin{tabular}{l p{10cm}}  
\toprule
\textbf{Feature} & \textbf{Description} \\ 
\midrule
\textbf{Atomic number} & Charge number of the atomic nucleus. \\ 
\textbf{Chirality} & Spatial configuration, e.g., CW/CCW or unspecified. \\ 
\textbf{Degree} & Number of neighboring atoms. \\ 
\textbf{Formal charge} & Integer electronic charge assigned to the atom. \\ 
\textbf{Number of Hs} & Number of bonded hydrogen atoms. \\ 
\textbf{Radical electrons} & Number of unpaired electrons. \\ 
\textbf{Hybridization} & Orbital hybrid type (e.g., sp, sp2, sp3). \\ 
\textbf{Aromaticity} & Whether the atom is part of an aromatic system. \\ 
\textbf{In ring} & Whether the atom belongs to a ring structure. \\ 
\bottomrule
\end{tabular}
\label{tab:graph_node_feature_table}
\end{table}

\subsection*{A2. Distributions of data}
\begin{figure}[h]
    \centering
    \includegraphics[width=0.53\linewidth]{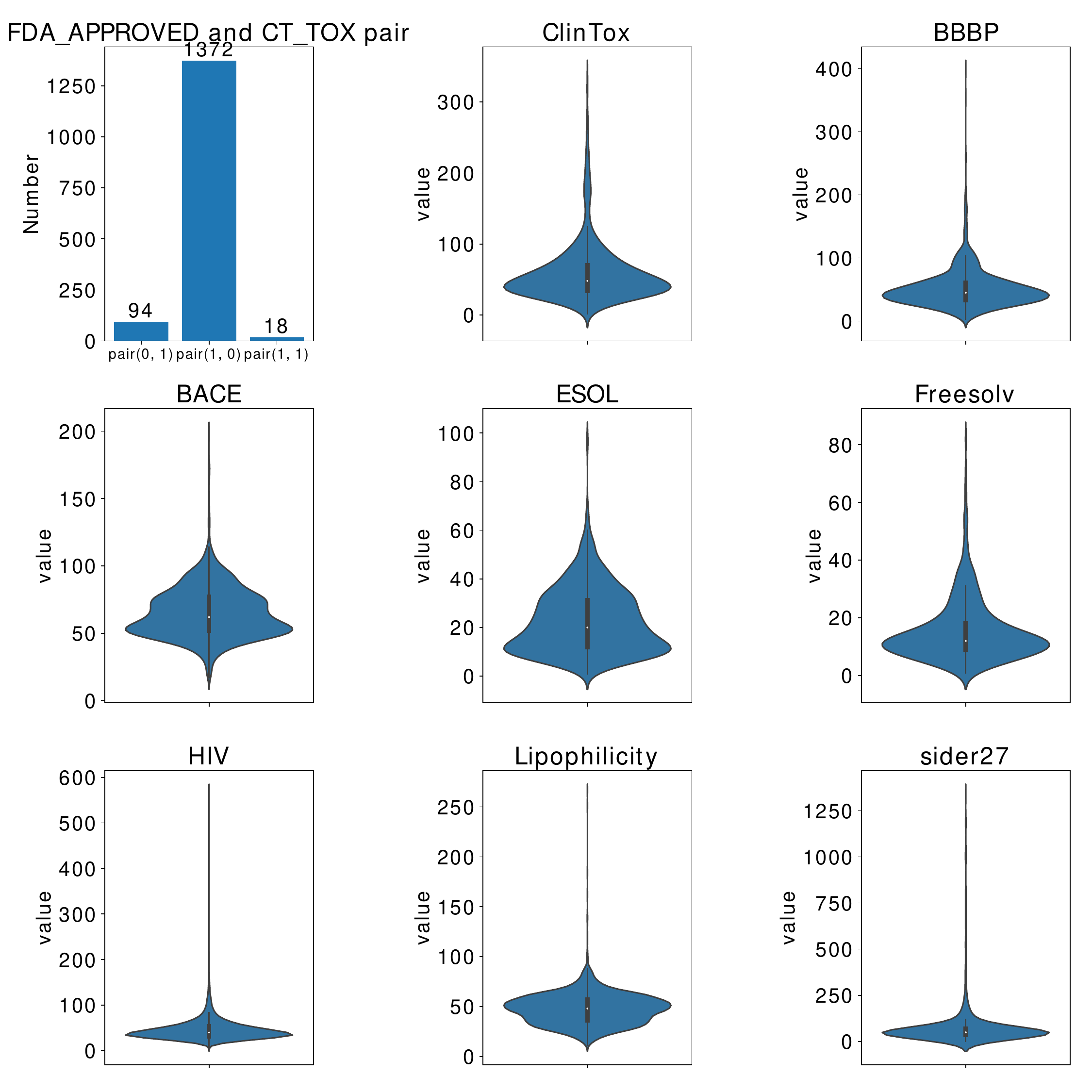} 
    \caption{The top-left sub-figure shows the distribution of the label combinations from the  ClinTox dataset. The rest sub-figures show the distributions of SMILES string lengths from all the eight datasets.}
    \label{fig:data_analysis}
\end{figure}
\subsection*{A3. Brief description of datasets}
Brief description of the datasets utilized in the experiments. More details can be found in the MoleculeNet Benchmark website \cite{wu2018moleculenet}.  
\begin{itemize}
\item \textbf{HIV:} This dataset involves experimentally measuring the abilities of molecules to inhibit HIV replication. 
\item \textbf{BACE:} It focuses on the qualitative binding ability of molecules as inhibitors of BACE-1 (Beta-secretase 1). 
\item \textbf{ClinTox:} It records whether the drug was approved by the FDA (Food and Drug Administration) and failed clinical trials for toxicity reasons.
\item \textbf{BBBP:} This dataset contains binary labels indicating blood-brain barrier penetration. 
\item \textbf{SIDER:} It comprises information on marketed drugs and their adverse drug reactions categorized into 27 system organ classes.
    \item \textbf{ESOL:} It contains water solubility data for molecules.
    \item \textbf{FreeSolv:} It records the hydration free energy of small molecules in water.
    \item \textbf{Lipophilicity:} It includes experimental results of octanol/water distribution coefficients.
\end{itemize}

\end{document}